\documentclass[a4paper,twoside]{article}
\usepackage{subcaption}
\usepackage{calc}
\usepackage{amssymb}
\usepackage{amstext}
\usepackage{amsmath}
\usepackage{amsthm}
\usepackage{multicol}
\usepackage{pslatex}
\usepackage{apalike}
\usepackage{algorithm2e}
\usepackage{float} 
\usepackage{graphicx}%
\usepackage{multirow}%
\usepackage{amsmath,amssymb,amsfonts}%
\usepackage{amsthm}%
\usepackage{mathrsfs}%
\usepackage[title]{appendix}%
\usepackage{xcolor}%
\usepackage{textcomp}%
\usepackage{manyfoot}%
\usepackage{booktabs}%
\usepackage{algpseudocode}%
\usepackage{listings}%
 \usepackage{url} 
 
\usepackage[bottom]{footmisc}
\usepackage{SCITEPRESS}     
\newcommand{\tb}[1]{\textcolor[rgb]{0.0,0.0,0.0}{#1}}    
\begin{document}

\title{\tb{Long-tailed Species Recognition in the NACTI Wildlife Dataset\vspace{-15pt}}}

\author{\tb{\authorname{Zehua Liu\sup{1}\orcidAuthor{0009-0004-5353-7028} and Tilo Burghardt\sup{1}\orcidAuthor{0000-0002-8506-012X}}}
\affiliation{\sup{1}School of Computer Science, University of Bristol, MVB Woodland Rd, BS8 1UB, Bristol, UK}
\email{[bw19062@bristol.ac.uk; tilo@cs.bris.ac.uk]}\vspace{-30pt}}


\abstract{\tb{As most ``in the wild'' data collections of the natural world, the North America Camera Trap Images (NACTI) dataset shows severe long-tailed class imbalance, noting that the largest `Head' class alone covers \(>\)50\% of the 3.7M images in the corpus.  Building on the PyTorch Wildlife model, we present a systematic study of Long-Tail Recognition methodologies for species recognition on the NACTI dataset covering experiments on various LTR loss functions plus LTR-sensitive regularisation. Our best configuration achieves 99.40\% Top-1 accuracy on our NACTI test data split, substantially improving over a 95.51\% baseline using standard cross-entropy with Adam. This also improves on  previously reported top performance in MLWIC2 at 96.8\% albeit using partly unpublished (potentially different) partitioning, optimiser, and evaluation protocols. To evaluate domain shifts (e.g. night-time captures, occlusion, motion-blur) towards other datasets we construct a Reduced-Bias Test set from the ENA-Detection dataset where our experimentally optimised long-tail enhanced model achieves leading 52.55\% accuracy (up from 51.20\% with WCE loss), demonstrating stronger generalisation capabilities under distribution shift. We document the consistent improvements of LTR-enhancing scheduler choices in this NACTI wildlife domain, particularly when in tandem with state-of-the-art LTR losses. We finally discuss qualitative and quantitative shortcomings that LTR methods cannot sufficiently address, including catastrophic breakdown for `Tail' classes under severe domain shift. For maximum reproducibility we publish all dataset splits, key code, and full network weights.\vspace{0pt}}}

\onecolumn 
\maketitle 



\begin{minipage}{\textwidth}
    \centering\vspace{-16pt}
    \includegraphics[width=392px,height=245px]{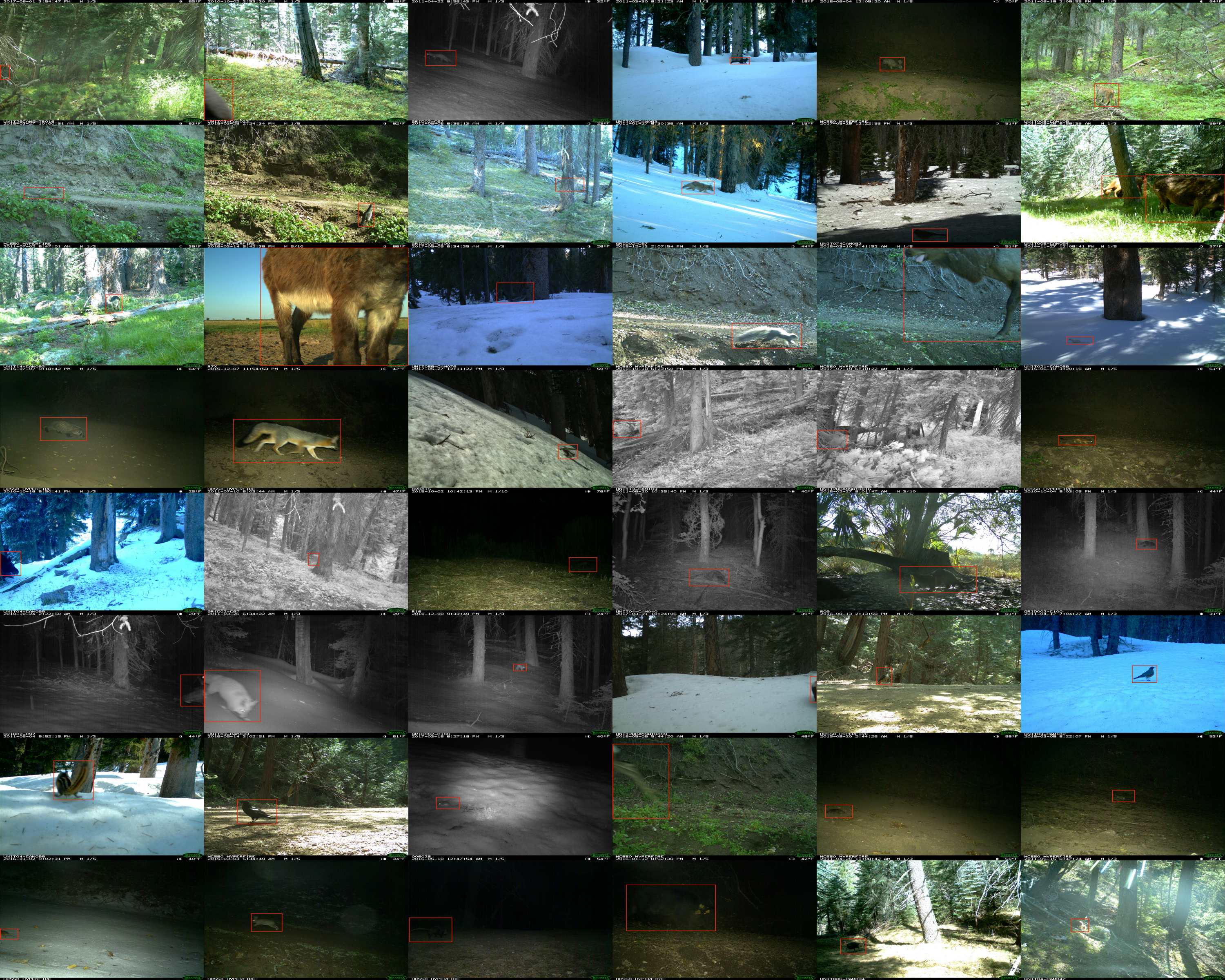}
    \captionof{figure}{\tb{{\textbf{Representative Samples from the NACTI Dataset.}  Shown is one representative image per species class from the overall 48 classes in total, excluding empty and vehicle frames. Note challenging occlusions, background clutter, motion blur, infrared shots, and small animal areas. Red bounding boxes indicate detections using~\texttt{MegaDetectorV6}, applied during the pre-processing stage. The NACTI dataset is long-tailed with over half of the samples belong to a single species (domestic cow). Following (Perrett et al.,2023), 5 classes form the head, 37 form the ‘long tail’, and 6 few-shot classes sit at the end of the species distribution.}}}
    \label{fig:overview_graphic}
\end{minipage}

\normalsize 
\setcounter{footnote}{0} 
\vfill

\twocolumn 
\section{\uppercase{Introduction}}
\label{sec:introduction}
\tb{\textbf{The Challenging NACTI Wildlife Dataset.} The North American Camera Trap Images (NACTI) dataset~\cite{NACTI2024} -- visually exemplified in Figure~\ref{fig:overview_graphic} -- contains 3.7M images from five U.S. sites covering 48 animal (sub)species. It presents a formidable challenge for automated wildlife recognition~\cite{kuhl2013animal,reynolds2025potential} due to extreme class imbalances. In NACTI, over half of the samples belong to a single species (domestic cow). Following~\cite{perrett2023useheadimprovinglongtail}, 5 classes form the head, 37 form the `long tail', and 6 few-shot classes sit at the end of the distribution where rare or endangered species are represented by often only a handful of examples. Furthermore, the dataset's untrimmed images captured under uncontrolled conditions (e.g., low illumination, motion blur) and its lack of bounding box or auxiliary metadata make fine-grained species ID challenging.}

\tb{\textbf{Limitations of Standard Vision Approaches.} Discussed issues are characteristic of most collected data from natural settings that conservationists and ecologists work with on a daily basis. While Convolutional Neural Networks (CNNs) as well as Transformer-based architectures can often achieve near human-level accuracy on images taken under ideal conditions~\cite{CNN_features} and~\cite{Dosovitskiy2020AnII}, their performance degrades sharply on long-tailed natural data. Tools like MegaDetector~\cite{beery2019efficient} and SpeciesNet~\cite{gadot2024crop} trained on many millions of images offer a strong foundation for general animal detection, but do not fully resolve the core issue of species-level classification under truly severe class imbalance. Consequently, standard models often fail decisively on minority classes, compromising the recognition of ecological animal encounters they are meant to provide for rare or endangered species that form the distribution tail~\cite{Chen2014}.}

\begin{figure*}[h]
    \centering
    \vspace{-17pt}
    \includegraphics[width=430px,height=190px]{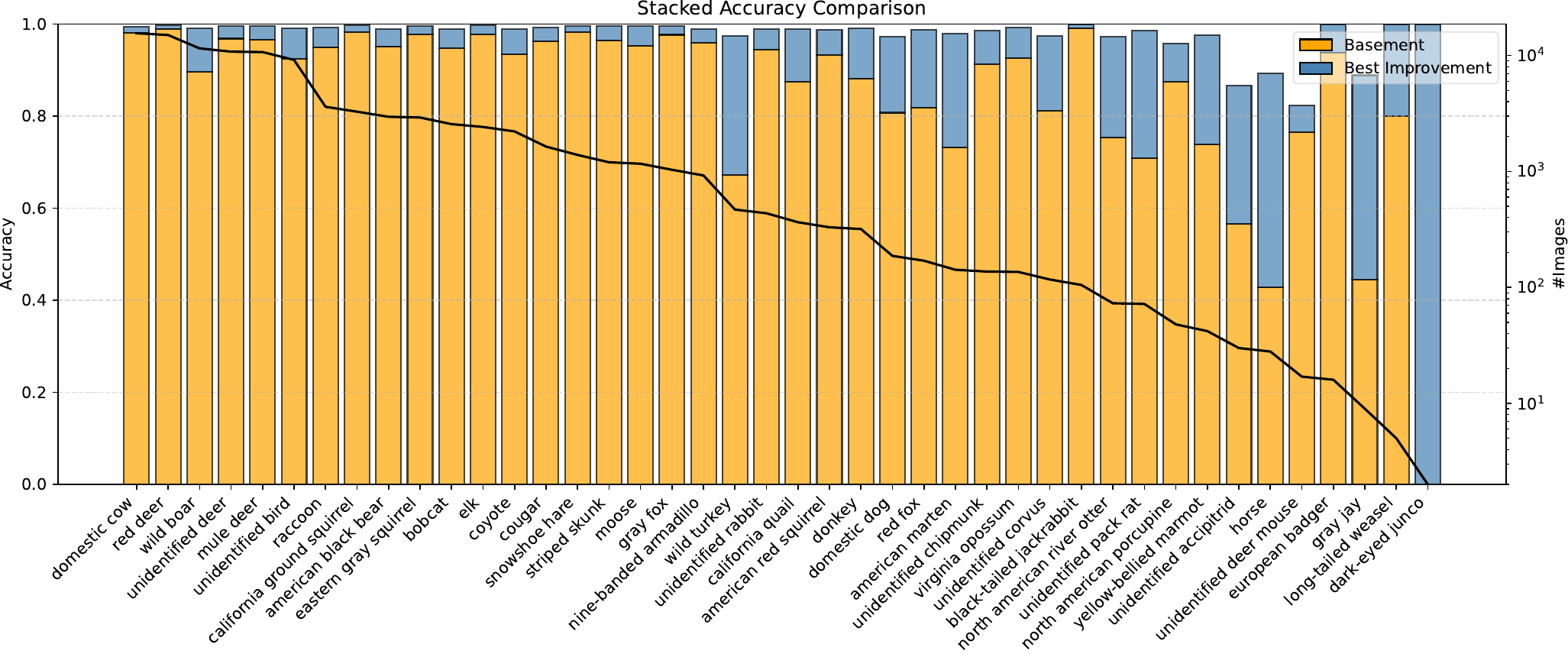}
    \captionof{figure}{\tb{{\textbf{LTR Performance Improvements across NACTI Species vs. Baseline.} Per-class Top-1 accuracies on the NACTI dataset comparing the baseline model (cross-entropy with Adam) against our best-performing configuration, which combines a Label-distribution-aware margin (LDAM) loss, an alternative AdamW optimiser, and a learning-rate scheduler. The results are reported on a 10\% random test split. The blue bars indicate classes with a positive improvement relative to the baseline.}}}\vspace{-5pt}
    \label{fig:baseline_vs_best}
\end{figure*}

\tb{\textbf{Our Approach and Contributions.} Towards tackling this critical issue in computer vision for wildlife monitoring~\cite{tuia2022perspectives}, our work presents a systematic, loss-centric study focused on improving minority class detection performance in the large-scale, real-world NACTI setting. We develop and present a scalable, multi-GPU pipeline to fine-tune state-of-the-art CNN architectures on the full NACTI corpus. Using this pipeline, we conduct a head-to-head comparison of Long-Tail Recognition (LTR) scheduling~\cite{alshammari2022longtailedrecognitionweightbalancing} and losses: focal loss~\cite{lin2018focallossdenseobject}, weighted cross-entropy~\cite{Cui2019ClassBalancedLoss}, and label-distribution-aware margin (LDAM) loss~\cite{cao2019learningimbalanceddatasetslabeldistributionaware}. Resulting from this analysis, Figure~\ref{fig:baseline_vs_best} summarises best model accuracy improvements over the baseline at class level. We also assess the generalisation potential of this model on a second camera-trap collection to quantify its robustness under distributional shift. Our paper contributions are threefold:}
\begin{enumerate}
  \item \tb{\textbf{First Systematic NACTI LTR Evaluation.} The first head-to-head evaluation of LTR-sensitive scheduling in tandem with focal, weighted cross-entropy, and LDAM LTR losses for large-scale wildlife classification, analysing trade-offs between minority-class recall and overall accuracy;}
  \item \tb{\textbf{Additional Out-of-Distribution (OOD) Robustness Analysis.} A cross-dataset assessment demonstrating that the best-performing loss maintains superior minority-class F$_1$ when species frequencies and imaging conditions change; and}  
\item \tb{\textbf{Reproducible Best Performance Classifiers.} Given limitations regarding the disclosure of details in much of prior art, we publish reproducible, fully open access multi-GPU networks that set the state-of-the-art on NACTI end-to-end and support easily comparable community research in future. We publish splits, code, and describe the approach in detail.}
\end{enumerate}

\vspace{-15pt}
\section{\uppercase{Key Related Work}}
\vspace{-7pt}
\subsection{\tb{Species Recognition in NACTI}}
\vspace{-5pt}

\tb{\textbf{General ResNet-based Approaches.} Prior art on species recognition in NACTI goes back several years, but most published works lack reproducibility and published descriptions/code that would allow for a fair and detailed comparison of approaches to illuminate the effect of chosen architectures regarding the long-tail distribution issue. Tabak et al.~\cite{tabak2022cameratrapdetector}, for instance, trained a ResNet-50 architecture on the NACTI dataset (70/30 split) using SGD with momentum -- however, the paper does not explicitly specify the loss function used; though one would assume a softmax-cross-entropy setting was applied. They reported \(\approx\)86\% accuracy on the validation set. Performance dropped significantly for under-represented species (recall \(<\)70\%), highlighting persistent long-tailed bias in camera-trap datasets.}

\tb{\textbf{Cross-Study MLWIC2 Project.} Tabak et al.~\cite{Tabak2020mlwic2} also trained MLWIC2, that is a ResNet-18 on $>3$M images from 18 North American studies (including NACTI), reporting 96.8\% Top-1 accuracy for the species model. To mitigate imbalance, the authors capped each class at 100{,}000 images. Despite these measures, rare species still showed weaker recall, indicating unresolved long-tail effects. For out-of-sample evaluations, reported Top-5 accuracy ranged from 65.2\% to 93.8\% depending on the dataset. We note, that the paper does neither specify optimiser nor loss function used.}

\tb{\textbf{Active Learning.} Norouzzadeh et al.~\cite{Norouzzadeh2021DeepActiveLearning} combined a ResNet-50 with active learning techniques to iteratively select informative, unlabelled images for annotation. After 30{,}000 queries, the classifier reached 93.2\% accuracy on NACTI. In their pipeline, the embedding network is trained either using softmax-cross-entropy or with triplet loss settings. However, the exact evaluation split is not specified and the optimiser used for these experiments is not explicitly specified in the paper either (SGD is discussed only as general background), so we treat it as unspecified.}

\tb{\textbf{Domain-aware DANAS Approach.} Jia et al.~\cite{jiadanas} introduced DANAS, a domain-aware neural architecture search tailored for camera-trap species classification. On the NACTI-a subset, DANAS showed that lightweight CNNs can match conventional performance baselines~(e.g., ResNet-18) within Top-1/Top-5 accuracy charts while using far fewer parameters and lower compute, enabling edge deployment at average accuracy \(\approx\)92.9\%. During the search phase, candidate networks were trained with AMSGrad~(learning rate = 0.005, batch size = 32) and guided by a customised, theoretically derived losses (Witch-of-Agnesi–based) that encourage high accuracy with few parameters. The optimiser used for the final training of the discovered DANAS networks is not explicitly stated, while baselines were stated and trained with Nesterov-SGD under a cosine loss regime~\cite{jiadanas}.}

\begin{figure*}[t]
  \centering\vspace{-17pt}
  \includegraphics[width=430px,height=155px]{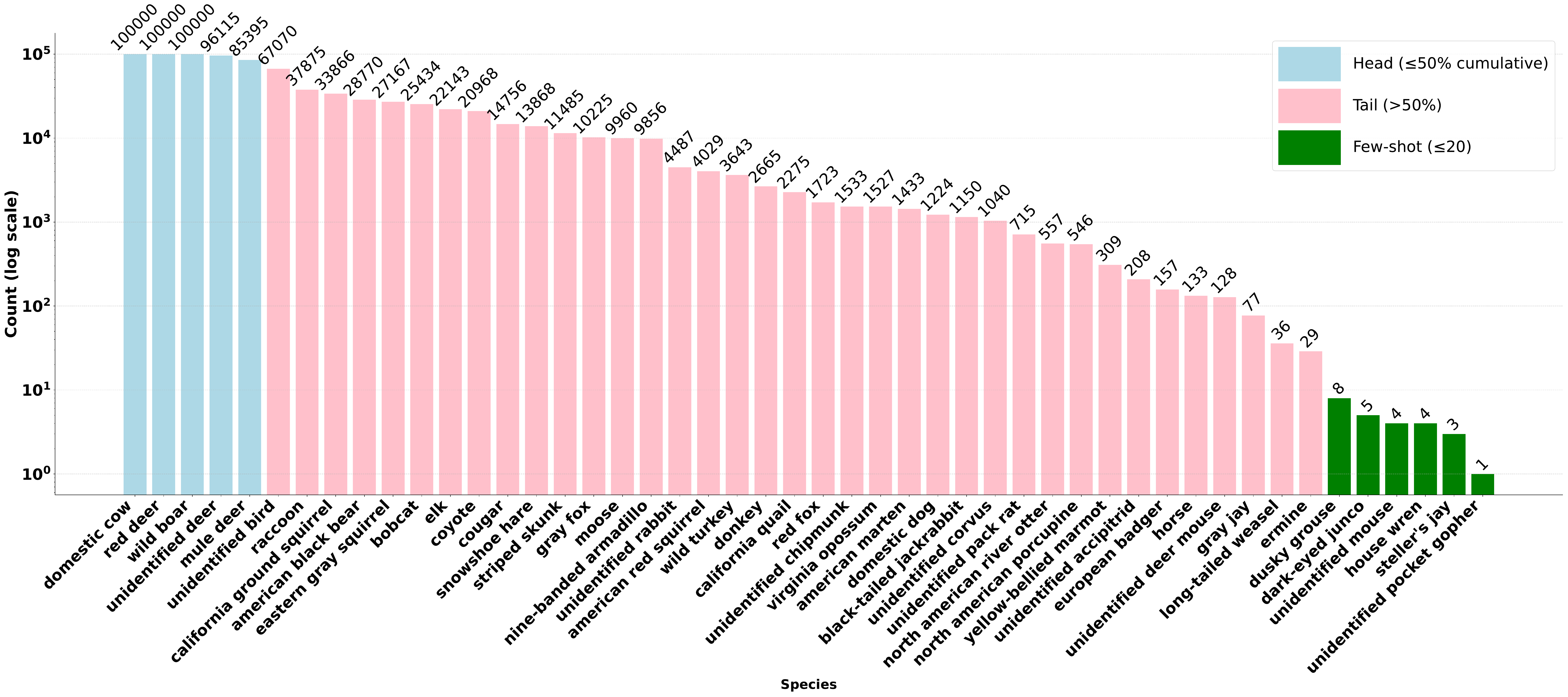}
  \caption{\tb{\textbf{NACTI Species Distribution after Sample Balancing.} Logarithmic visualisation of the cardinality of all 48 (sub)species classes of NACTI after capping classes at 100k. Colouration of the class distribution after~\cite{perrett2023useheadimprovinglongtail} breaking the distribution into `Head' (blue), `Tail' (pink), and `Few-shot' species.}}\vspace{-5pt}
  \label{fig:balanced_distribution}
\end{figure*}

\vspace{-5pt}
\subsection{\tb{Regularisers and Losses for LTR}}
\vspace{-5pt}
\tb{\textbf{Scheduling for LTR.} Addressing class imbalance, a central challenge in real-world data~\cite{zhang2023deeplongtailedlearningsurvey}, has led to various strategies. Recent study~\cite{alshammari2022longtailedrecognitionweightbalancing} have highlighted the importance of weight decay in mitigating classifier bias towards dominant classes, which is a key challenge in long-tailed recognition tasks. Alshammari et al.~\cite{alshammari2022longtailedrecognitionweightbalancing} showed that models trained without weight decay tend to assign disproportionately large classifier weights to frequent classes, resulting in poor generalisation for rare classes. By carefully tuning weight decay parameters, they significantly improved recognition performance on long-tailed datasets. Unlike conventional L1 or L2 regularisation, their method effectively balanced per-class classifier weights, thereby preventing common species from dominating learning. Furthermore, their study demonstrated that simple weight decay tuning outperformed many sophisticated LTR methods, suggesting that parameter regularisation should be of primary consideration when addressing class imbalance.}

\tb{\textbf{Decoupled Learning Strategies.} A seminal work by Kang et al.~\cite{kang2020decouplingrepresentationclassifierlongtailed} proposed decoupling feature representation learning from classifier learning. They observed that standard instance-balanced sampling yields high-quality representations, while the classifier benefits from a class-balanced perspective, e.g. re-weighting or re-sampling. Their work also introduced LDAM-DRW, which combines the LDAM loss with a \textit{Deferred Re-weighting (DRW) schedule}, delaying the application of class-balancing until later in training. This concept strongly aligns with our findings, where combining the LDAM loss with a learning rate (LR) scheduler (ReduceLROnPlateau) yielded best results (see Section~\ref{ResLab} for details).}



\tb{\textbf{Problem-specific LTR Loss Concepts.} To systematically address the severe class imbalance in the NACTI dataset, we implemented, compared, and evaluated three state-of-the-art loss functions that modify the standard cross-entropy in complementary ways. Our goal was to isolate the most effective strategy for improving performance on minority classes within a large-scale, real-world setting.}

\tb{\textbf{Focal Loss.} Focal loss~\cite{lin2018focallossdenseobject} reduces the weight of easily classified samples, encouraging the model to focus more on hard-to-classify examples. This improves performance in classes with fewer samples.}

\tb{\textbf{Weighted Cross-Entropy~(WCE) Loss.} The WCE concept~\cite{Cui2019ClassBalancedLoss} addresses class imbalance by assigning greater weights to minority classes, thereby encouraging the model to focus more on under-represented categories during training and improving their recognition rates. However, this can lead to overfitting, causing the model to prioritise minority classes at the expense of overall performance.}

\tb{\textbf{Label-Distribution-Aware Margin (LDAM) Loss.} LDAM~\cite{cao2019learningimbalanceddatasetslabeldistributionaware} addresses class imbalance by encouraging larger decision margins for minority classes. It modifies the standard cross-entropy loss by subtracting a class-dependent margin from the logit of the true class. This formulation gently penalises confident predictions for frequent classes while imposing stronger regularisation on minority classes. As a result, LDAM improves generalisation across the label distribution, especially for under-represented classes. It can also be combined with re-weighting or re-sampling strategies for further performance gains.}



\vspace{-14pt}
\section{\uppercase{Data Preparation}}
\vspace{-7pt}
\subsection{\tb{Dataset Statistics}}\vspace{-5pt}
\label{Section:Dataset}
\tb{\textbf{The NACTI Dataset.} Published in the LILA Dataset collection~\cite{NACTI2024}, NACTI contains 3.7M camera-trap images from spanning 48 (sub)species. Figure~\ref{fig:overview_graphic} displays representative images from each species class, illustrating the wide variability in appearance, lighting, and environmental conditions. These samples exemplify common visual challenges encountered in camera trap imagery, including occlusions, background clutter, motion blur, and significant class imbalance. Such factors complicate the training process significantly particularly for under-represented classes where few images are available to capture species-distinctive features.}

\tb{\textbf{Species Distribution Details.} The class distribution is extremely long-tailed; \textit{domestic cow} alone contributes 2,109,009 images (approx. $54\%$ of all samples). To reduce overfitting and speed up training, we cap each class at 100k images, following Tabak \textit{etal.}~\cite{Tabak2020mlwic2}. This yields 816,495 balanced samples (Fig.~\ref{fig:balanced_distribution}) while preserving all 48 classes. The visualisation approach follows that of Perrett et al.~\cite{perrett2023useheadimprovinglongtail}, where the dataset is divided into three groups: `Head' classes (blue) contain $>50\%$ of total samples, `Tail' classes (pink) cover most species, while `Few-shot' classes (green) include those with $<20$ samples. }

\begin{figure*}[t]\vspace{-6pt}
    \centering 
    \begin{subfigure}[b]{0.49\textwidth}
        \centering
        \includegraphics[width=\linewidth]{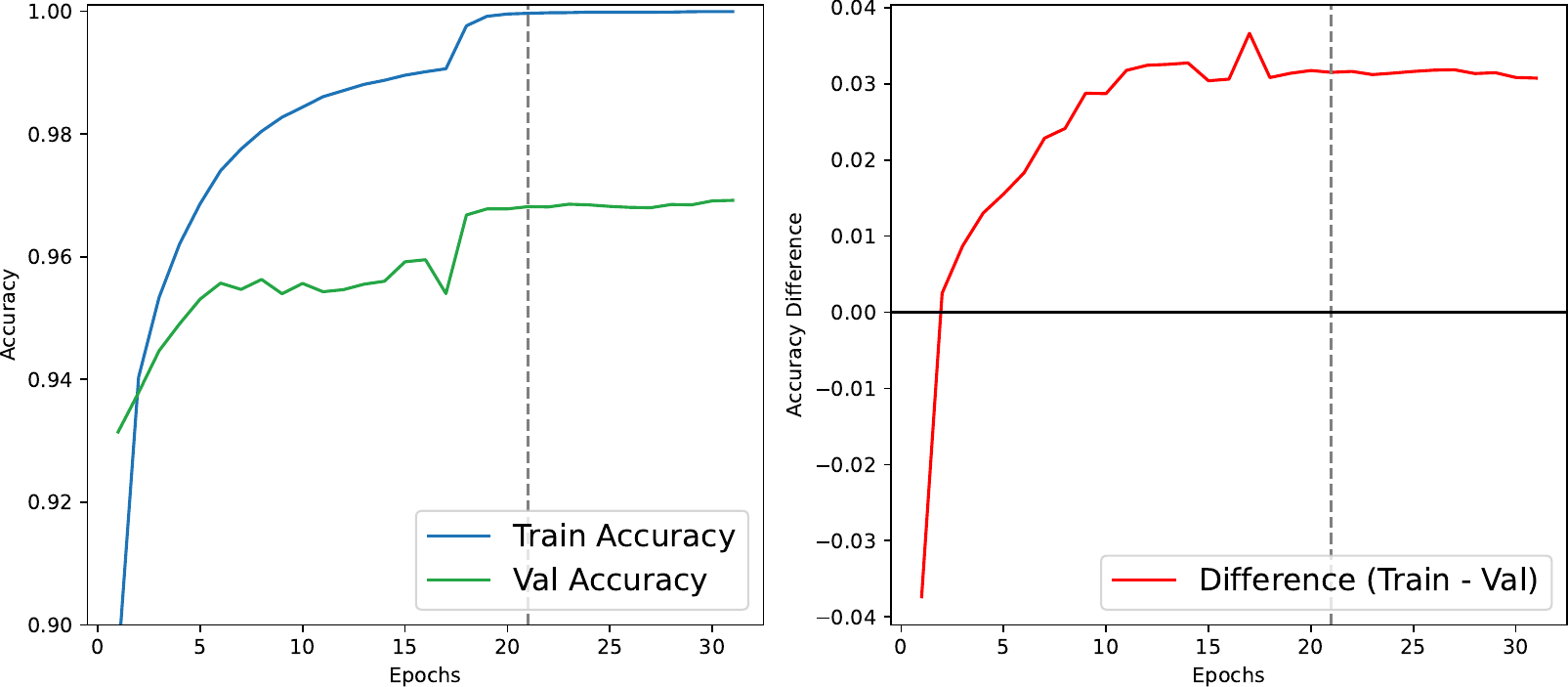}
        \caption{Accuracy Metrics}
        \label{fig:acc_curves}
    \end{subfigure}
    \hfill 
    \begin{subfigure}[b]{0.49\textwidth}
        \centering
        \includegraphics[width=\linewidth]{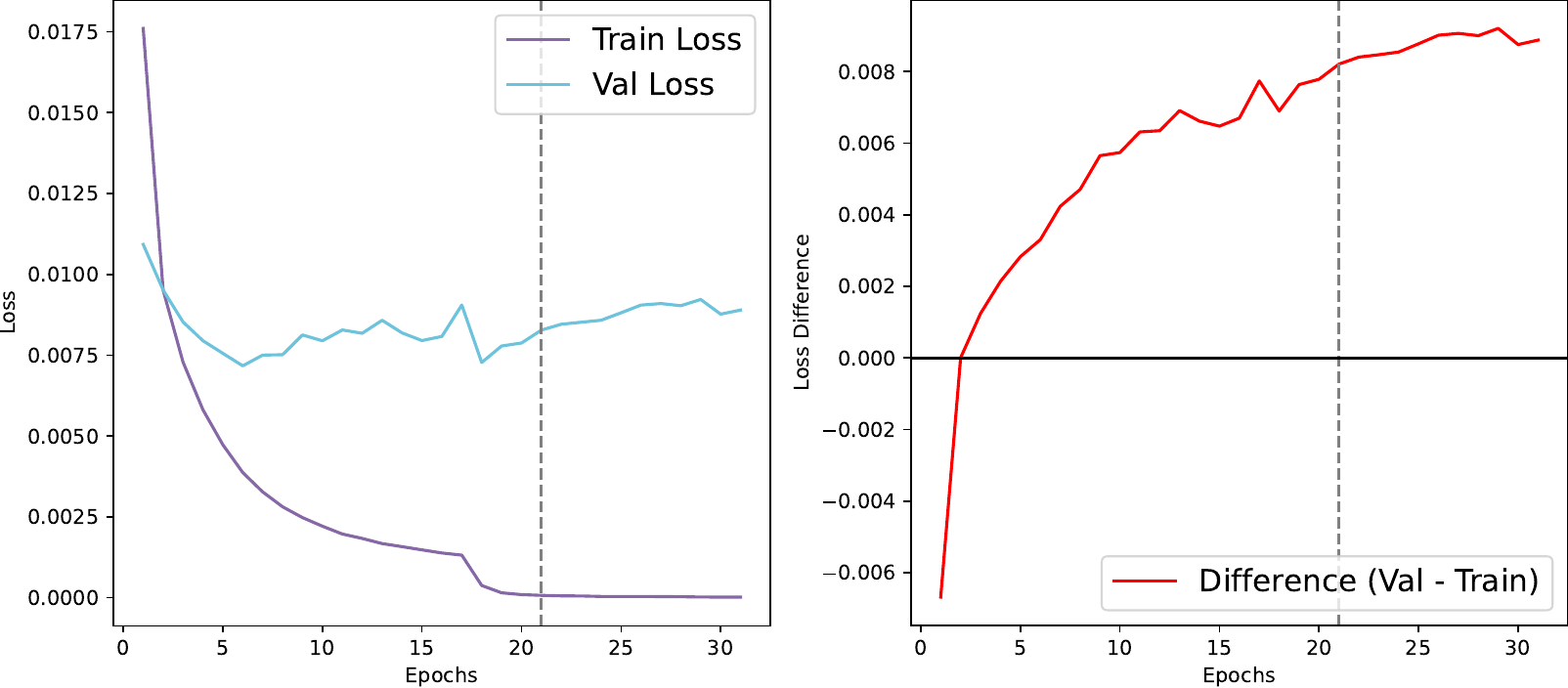}
        \caption{Loss Metrics}
        \label{fig:loss_curves}
    \end{subfigure}\vspace{6pt}
    \caption{\tb{\textbf{Training Dynamics.} Column \textbf{(a)} reports accuracy whilst column \textbf{(b)} reports on loss on one of the training runs with scheduling. Within each column, the left plot shows train and validation, right plots show the train-validation gap. Note the dashed lines marking  early-stopping point.}}\vspace{-6pt}
    \label{fig:training_dynamics}
\end{figure*}

\vspace{-7pt}
\subsection{Data Preprocessing}\vspace{-8pt}
\tb{Our workflow converts the raw image NACTI archive into an analysis-ready dataset in three stages.}

\tb{\textbf{(i) Automatic Animal Localisation.}  
 All images are first processed with~\texttt{MegaDetectorV6} to obtain bounding boxes and confidence scores. The resulting JSON files are merged with the original metadata such that each record includes both species labels and coordinates for the detected animal region.}

\tb{\textbf{(ii) Class-Balanced Resampling.}  
Following Tabak~\emph{et al.}, we cap every category at 100,000 samples, discard empty/vehicle frames, and retain all 48 animal classes. This produces 816,495 images with a more balanced distribution profile shown in Figure~\ref{fig:balanced_distribution}.}

\tb{\textbf{(iii) Split Strategy and Reduced-Bias Evaluation.}  
The balanced pool is split \textit{80/10/10} (train/validation/test) with a fixed seed, acknowledging that visually similar burst information may still be spread across subsets in this setting. To gauge towards bias reduction and better generalisation we also construct and publish a \emph{reduced-bias test set} using the ENA Dataset~\cite{ena2024}, filtering to the nine classes shared with NACTI and removing any temporal/spatial overlap. All nine fall in the distribution `Tail’, providing a stringent domain-shift benchmark.}

\tb{\textbf{Final Pre-processing Steps.}  Images are resized to $256\!\times\!256$ px as per Tabak~\emph{et al.}, converted to tensors, and normalised with \(\mu=[0.485,0.456,0.406]\), \(\sigma=[0.229,0.224,0.225]\) to stabilise optimisation. Each training sample is finally stored as an information tuple~\texttt{(image, \{bbox, label, conf\})}.}

\vspace{-14pt}
\section{\uppercase{Experiments}}\vspace{-7pt}
\subsection{Baseline Model}\vspace{-7pt}
\tb{\textbf{Hardware Specification.} All training is conducted utilising 11 NVIDIA Tesla P100 GPUs~(Pascal architecture) to efficiently distribute computation across multiple devices and reduce training time. All inference was performed locally on an NVIDIA GeForce RTX 3090 GPU~(Ampere architecture).}

\begin{table}[htbp]\vspace{8pt}
    \caption{\tb{\textbf{NACTI Test Accuracy (LTR Methods vs. Baselines)} Comparison of (Overall/`Tail') accuracy on the biased NACTI test set across all configurations. Note that `Tail' performance stats are saturated with the top accuracy of 99.22\% is shared by three configurations.}}
    \centering
    \scriptsize
    \begin{tabular}{|l|c|c|}
    \hline
    \textbf{LOSS + OPTIMISER} & \textbf{BASELINES} & \textbf{LTR LR}\\
        \textbf{CONFIGURATION} & (NO SCHEDULER) & \textbf{ SCHEDULER} \\
    \hline
     \textbf{\textit{BASELINES}}   &  &  \\
    Cross-Entropy + Adam   & 95.51\% / 94.30\% & Not Tested \\
    Cross-Entropy + AdamW  & 96.77\% / 95.64\% & 99.34\% / \textbf{99.22\%} \\
        \hline
    \textbf{\textit{LTR LOSSES}}   &  &  \\
    Focal Loss + AdamW  & 96.65\% / 95.96\% & 99.31\% / 99.20\%  \\
    WCE + AdamW & 96.69\% / 96.16\% & 99.35\% / \textbf{99.22\%} \\
    LDAM Loss + AdamW & 96.71\% / 96.17\% & \textbf{99.40\%} / \textbf{99.22\%} \\
    \hline
    \end{tabular}
    \label{tab:configuration_comparison_biased_tail}
\end{table}

\tb{\textbf{Full Network Training Details.} All experiments fine-tune the ResNet-50~\cite{he2016deep} backbone of \texttt{AI4GAmazonRainforest}~\cite{hernandez2024pytorchwildlife}  with its standard pre-trained weights, using AdamW~\cite{Loshchilov2017DecoupledWD} (weight-decay $10^{-2}$, initial learning rate $10^{-4}$) and a \texttt{ReduceLROnPlateau} scheduler that lowers the learning rate by factor 0.1 when validation metrics stagnate. To avoid premature termination or overfitting caused by fluctuations in validation performance we employ early stopping (see Figure~\ref{fig:training_dynamics}) with patience 10 and check-pointing, dynamically selecting the best model based on validation recall~(see published code).}

\vspace{-5pt}\subsection{Experimental Results}\vspace{-5pt}\label{ResLab}
\tb{\textbf{Scheduler Impact on NACTI Results.} Table~\ref{tab:configuration_comparison_biased_tail} compares various combinations of LTR-sensitive learning rate scheduling and LTR loss functions+optimisers against baselines, evaluated in terms of both overall accuracy and performance on the `Tail' classes. Introducing a learning rate scheduler from PyTorch~\cite{pytorch} resulted in substantial performance gains across all configurations, with overall accuracy consistently exceeding 99\% on the biased test set. The overall best result was achieved by combining LDAM loss with the AdamW optimiser and the \texttt{ReduceLROnPlateau} scheduler, yielding an accuracy of 99.40\%—an improvement of approximately 2.66 percentage points over the same configuration without a scheduler (96.74\%).}

\tb{\textbf{NACTI Tail Results.} A similar trend was observed for the `Tail' classes: accuracy increased from 96.17\% to 99.22\%, representing a relative improvement of 3.05\%. Although all configurations achieved comparable performance on the `Tail' classes following scheduler integration, the LDAM + AdamW combination attained the highest overall accuracy and is henceforth referred to as \textit{best (LTR) configuration}.}

    

\tb{\textbf{Reduced-Bias Test Set Results.} All configurations are additionally evaluated on the independently constructed reduced-bias test set for assessing generalisation performance under distribution shift. Table~\ref{tab:configuration_comparison_unbiased} summarises this task showing expected, substantial accuracy drops relative to Table~\ref{tab:configuration_comparison_biased_tail}. While all configurations perform markedly worse in this OOD setting, using a learning rate scheduler yields consistent improvements across all loss functions. Among them, the LDAM loss combined with AdamW and a scheduler—our best configuration on the biased test set—achieves the highest accuracy (52.55\%), demonstrating its robustness under both class imbalance and domain shift.}


\vspace{4pt}
\begin{table}[h]
    \caption{\tb{\textbf{Reduced-Bias Test Dataset Accuracy.} Note that learning rate scheduling consistently improves performance and that the LDAM-Scheduler combination shows best performance in this OOD task.}}
    \centering
    \scriptsize
    \begin{tabular}{|l|c|c|}
        \hline
    \textbf{LOSS + OPTIMISER} & \textbf{BASELINES} & \textbf{LTR LR}\\
        \textbf{CONFIGURATION} & (NO SCHEDULER) & \textbf{ SCHEDULER} \\
        \hline
             \textbf{\textit{BASELINES}}   &  &  \\
        Cross-Entropy with Adam   & 39.48\% & Not Tested \\
        Cross-Entropy with AdamW  & 38.27\% & 48.27\% \\
                \hline
        \textbf{\textit{LTR LOSSES}}   &  &  \\           
        Focal Loss with AdamW  & 37.62\% & 49.78\%   \\
        WCE with AdamW & \textbf{45.03\%} & 51.20\%\\
        LDAM Loss with AdamW & 38.77\% & \textbf{52.55\%} \\
        \hline
    \end{tabular}
    \label{tab:configuration_comparison_unbiased}
\end{table}\vspace{-4pt}
\begin{table}[htbp]
    \caption{\tb{\textbf{Per-class Accuracy and Prevalence.} Results are shown for the Reduced-Bias Test Set under our best configuration.}}
    \centering
    \resizebox{\linewidth}{!}{%
    \begin{tabular}{|p{3.4cm} | r | r|}
    \hline
    \textbf{CLASS} & \textbf{ACCURACY (\%)} & \textbf{PREVALENCE}\\
    \hline
    virginia opossum & 65.79\% & 725  \\
    bobcat & 61.26\% & 333 \\
    american black bear & 77.05\% & 475  \\
    eastern gray squirrel & 45.51\% & 312 \\
    wild turkey & 0.00\% & 427  \\
    striped skunk & 21.55\% & 297 \\
    red fox & 54.72\% & 413 \\
    horse & 0.00\% & 63 \\
    coyote & 87.79\% & 344  \\
    \hline
    \end{tabular}%
    }
    \label{tab:nonzero_accuracy_with_horse}
\end{table}
\vspace{-4pt}
\vspace{-22pt}
\section{\uppercase{Discussion}}\vspace{-8pt}
\subsection{Generalisation Considerations}\vspace{-8pt}
\tb{\textbf{Testing Bias and Performance Saturation.}  As discussed in Section~\ref{Section:Dataset}, the standard NACTI Test Set -- using the standard random selection process for testing NACTI so far -- contains numerous visually sim-}


\noindent
\tb{ilar frames originating from camera-trap burst capture, making it especially vulnerable to biased testing. As a result, test accuracy fails to reflect generalisability and leading to saturated 99\%+ performances. However, the model's ability to generalise is shown on the reduced-bias test set where different LTR loss functions can offer minor accuracy improvements as seen in Table~\ref{tab:configuration_comparison_unbiased}. The dramatic performance gap between biased and reduced-bias test sets underscores the need for model-external OOD evaluation particularly for wildlife data like NACTI where significant context changes are part of environmental variability.}

\begin{figure}[t]
  \centering
  \vspace{-11pt}\includegraphics[width=0.99\linewidth]{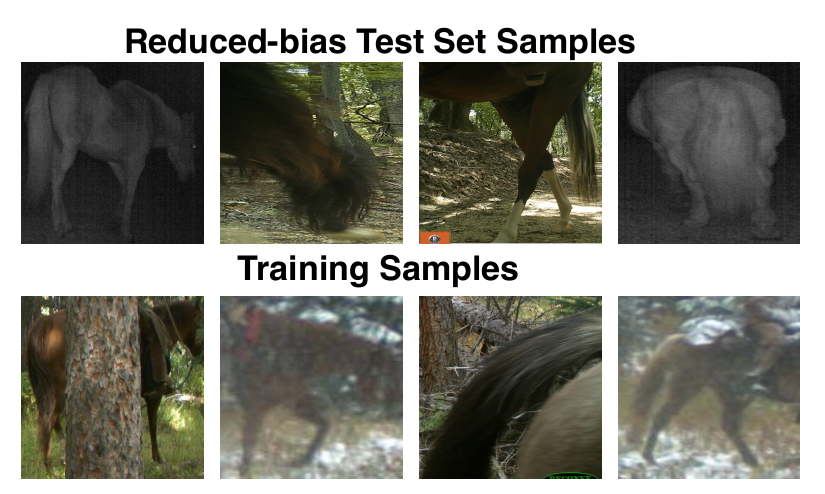}
  \caption{\tb{\textbf{Qualitative Example regarding Domain Shift for `Horse'.} Train-test contrast of some samples from the \textit{horse} class where we observed catastrophic performance breakdown. Training images are often blurry or occluded, whereas reduced-bias test images involve night scenes, partial views and unusual poses.}}\vspace{5pt}
  \label{fig:horse_examples}
\end{figure}

\tb{\textbf{Per-Class Generalisation Analysis.}
To understand the generalisation behaviour of the best-performing model further on the Reduced-Bias Dataset, per-class accuracies for the 9 classes in this dataset are reported in Table~\ref{tab:nonzero_accuracy_with_horse}. Note that all listed species belong to the `Tail' of the training distribution. While the model demonstrates relatively strong performance on classes such as \textit{coyote} and \textit{american black bear}, accuracy varies considerably across species. Notably, some classes like \textit{wild turkey} and \textit{horse} receive 0\% accuracy. These results underscore the difficulty of long-tailed generalisation under distribution shift. These findings suggest that while optimisation strategies such as LTR loss functions and learning rate schedules can improve overall accuracy, operation on rare classes may underperform catastrophically under distribution shift. Persistently low or zero accuracy on certain species across all configurations indicates that representational limitations play a more significant role unrecoverable by means of current learning alterations, potentially requiring a data-centric approach.}

\begin{figure}[t]
  \centering
  \vspace{-11pt}
  \includegraphics[width=0.99\linewidth]{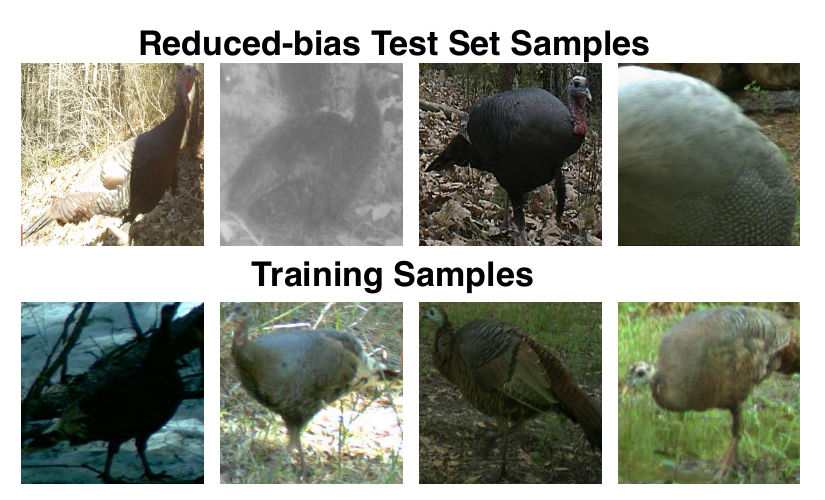}
  \caption{\tb{\textbf{Qualitative Examples regarding Domain Shift for `Wild Turkey'.} 
  Test images show poorer illumination, heavier occlusion and greater background clutter, widening the domain gap.}}
  \label{fig:wild_turkey_examples}
\end{figure}

\vspace{-4pt}
\subsection{Failure Mode Analysis}\vspace{-7pt}
\label{sec:consistent_failures}

\tb{\textbf{Predictive Collapse for Tail Classes.} Two classes -- \textit{horse} (\(n=63\)) and \textit{wild turkey} (\(n=148\)) --  consistently rank among the lowest-performing classes in the biased NACTI test set across \emph{all} configurations and accuracy for both classes collapses to approximate 0\% in the Reduced-Bias Test Set.  Tables~\ref{tab:horse_preds} exemplifies this collapse quantitatively. The model struggles to learn fine-grained distinctions between semantically or morphologically similar species when the training data does not contain sufficient examples of classes that actually contain species-determining features. This poses an ill-posed task of recognition of species who have little to no definition during the training phase in the first place. The consistent yet unsurprising misclassification into frequent categories confirms a strong prediction bias toward `Head' classes, reflecting the model’s defaulting to dominating features learned from majority categories.} 

\tb{\textbf{Qualitative Evidence.} Figures~\ref{fig:horse_examples} and~\ref{fig:wild_turkey_examples} juxtapose representative images from the NACTI training set and the Reduced-Bias Test Set. Training samples are often blurry or partly occluded, whereas test images often contain only partial views or nocturnal infrared captures. Two interacting factors contribute to these systematic errors:}

\begin{enumerate}
    \item \tb{\textbf{Class-Imbalance Bias.} Misclassifications of severely under-represented classes in the training distribution occur due to network defaulting to dominant visual cues learned from the majority classes, yielding skewed posteriors that rely on class distribution priors rather than class-indicative image data.}
    \item \tb{\textbf{Domain Shift.} Compared with training images, our Reduced-Bias Test shots exhibit different lighting, more partial occlusion, night scenes and stronger motion blur overall.  These discrepancies widen the representational gap and can lead to total detection collapse~(see Table~\ref{tab:horse_preds}).}
\end{enumerate}

\noindent
\tb{\textbf{{Partial View Considerations.}} For many tail classes including \textit{horse}, the network almost invariably predicts dominant classes, for \textit{horse} mainly \textit{red deer} or \textit{domestic cow} whenever a full body is not visible. This not only reflects the model’s bias towards the most frequent `Head’ classes under uncertainty, but points particularly to a problem of `unseen' views. This observation points in the direction of future work to analyse data augmentation and the simulation of partial view detection and related generative data strategies to compliment LTR scheduling and training methodologies.}


\vspace{-15pt}
\section{\uppercase{Conclusions}}\vspace{-6pt}
\label{sec:conclusion}

\tb{\textbf{Reproducible NACTI SOTA Performance.} This paper delivers a reproducible improvement in long-tailed species recognition for the NACTI dataset. To make these capabilities accessible to the community, we publish the end-to-end pipeline that processes raw NACTI frames and fine-tunes a ResNet-50-based architecture with LTR losses under Adam/AdamW optimisers. We confirm experimentally that a scheduled LDAM/AdamW configuration will yield overall 99.4\% Top-1 accuracy, setting a new reference point for NACTI wildlife species recognition under long-tailed conditions. We publish full code, weights, and complete data split information for reproducibility.}  
\vspace{5pt}
\begin{table}[h]
  \centering
  \caption{\tb{\textbf{Collapse of Classification for \textit{`horse'}.} Distribution of class recognitions (all false) of best performing LTR model in the Reduced-Bias Test Set confirms catastrophic collapse of classification with the  classifier defaulting to head classes even under maximal use of LTR adjustments.}}
  \label{tab:horse_preds}
  \begin{tabular}{|l|r|r|}
    \hline
    \textbf{PREDICTED CLS}       & \textbf{COUNT} & \textbf{PERCENT} (\%) \\ 
    \hline
    domestic cow           & 28 & 44.4 \\
    red deer               & 19 & 30.2 \\
    coyote                 &  6 &  9.5 \\
    cougar                 &  4 &  6.3 \\
    american black bear    &  3 &  4.8 \\
    unidentified deer      &  1 &  1.6 \\
    mule deer              &  1 &  1.6 \\
    unidentified bird      &  1 &  1.6 \\
    \hline
  \end{tabular}
\end{table}\vspace{5pt}

\tb{\textbf{OOD Analysis and Recommendation.}  We also provide results on domain shifted classifier application for the \textsc{ENA-Detection} Dataset, highlighting both the generalisation improvements and the large remaining generalisation gap in such wildlife OOD experiments which current technology is not able to overcome without further training information. For all tested settings, we confirm consistent improvements of LTR-enhancing scheduling in and out of domain demonstrating best generalisation capabilities when scheduling is combined with LDAM losses. Thus, given these insights on NACTI in context with previous work~\cite{alshammari2022longtailedrecognitionweightbalancing} we recommend this setup as a basic starting point for addressing LTR tasks in wildlife monitoring. We hope that the publication of a reproducible SOTA pipeline for NACTI encourages further research in the evolving and important field of computer vision for wildlife monitoring.}




\bibliographystyle{apalike}



\end{document}